\documentclass{article}
\usepackage{amsmath,epsfig}
								
\usepackage{caption}
\usepackage{amssymb}
\usepackage{multirow}
\usepackage{mathtools}
\usepackage{siunitx}
\usepackage[preprint]{spconfa4}

\let\OLDthebibliography\thebibliography
\renewcommand\thebibliography[1]{
  \OLDthebibliography{#1}
  \setlength{\parskip}{0pt}
  \setlength{\itemsep}{0pt plus 0.3ex}
}

\begin{document}\sloppy

\title{Supervised Video Summarization via Multiple Feature Sets with Parallel Attention}

\name{Junaid Ahmed Ghauri$^1$, Sherzod Hakimov$^1$, and Ralph Ewerth$^{1,2}$}
\address{$^1$ TIB -- Leibniz Information Centre for Science and Technology, Hannover, Germany
\\
$^2$ L3S Research Center, Leibniz University Hannover, Germany\\
\{junaid.ghauri, sherzod.hakimov, ralph.ewerth\}@tib.eu}

\maketitle

\begin{abstract}
The assignment of importance scores to particular frames or (short) segments in a video is crucial for summarization, but also a difficult task. Previous work utilizes only one source of visual features. 
In this paper, we suggest a novel model architecture that combines three feature sets for visual content and motion to predict importance scores.
The proposed architecture utilizes an attention mechanism before fusing motion features and features representing the (static) visual content, i.e., derived from an image classification model.
Comprehensive experimental evaluations are reported for two well-known datasets, SumMe and TVSum. 
In this context, we identify methodological issues on how previous work used these benchmark datasets, and present a fair evaluation scheme with appropriate data splits that can be used in future work. 
When using static and motion features with parallel attention mechanism, we improve state-of-the-art results for SumMe, while being on par with the state of the art for the other dataset.

\end{abstract}
\section{Introduction}
\label{sec:intro}

In the current information age, the enormous amount of available informative or entertaining multimedia content has increased the need for methods to detect important and thus relevant content. The number of available videos is also growing rapidly~\cite{DBLP:conf/cvpr/SongVSJ15,DBLP:conf/eccv/ZhangCSG16} and the velocity highlights that the detection of important video segments is an essential and crucial task for the field of computer vision. Every minute, hundreds~\cite{DBLP:conf/eccv/ZhangCSG16,DBLP:conf/eccv/ZhangGS18} or even a thousand hours of videos are being uploaded to video or social media platforms. 
It can be observed that skipping forward to a desired or interesting part of a video is widespread and this behavior is also denoted as "skim through"~\cite{DBLP:conf/eccv/PotapovDHS14}.
No doubt this is subjective behavior but similar behavior of various humans can highlight the importance of a particular video segment. Overall, it is obvious that the fine-grained identification of important segments in a video is an important task.
Video summarization can be defined as the conversion of a (potentially long) video into a shorter video that contains essential segments and thus allows a viewer to understand the video content. 
One of the main challenges is to identify and select important frames and segments in the original video that can be reused in the video summary.
The generation of summaries in the form of selected frames from a video is useful when the main task is to get shorter videos with possible important parts. 
Many methods apply video segmentation as an initial pre-processing step for video summarization~\cite{DBLP:conf/cvpr/SongVSJ15,DBLP:conf/eccv/ZhangCSG16,DBLP:conf/eccv/GygliGRG14,DBLP:conf/mm/Wang000FT19}. 
For example, fine-grained approaches at the video segment level of a second can open doors to potential  \emph{applications} such as live video stream from surveillance, learning, or the entertainment sector. 
Although the problem of assigning importance scores to frames for video summarization has been studied before ~\cite{DBLP:conf/eccv/ZhangCSG16,DBLP:conf/eccv/GygliGRG14, DBLP:conf/mm/Wang000FT19,DBLP:conf/eccv/PotapovDHS14,DBLP:conf/nips/ShiCWYWW15, DBLP:conf/nips/GongCGS14}, little attention has been paid to the impact of core system components, that is the role of different types of visual features and how to combine them. In this regard, most previous work just relied on content-based features for image classification~\cite{DBLP:conf/eccv/ZhangGS18,DBLP:journals/corr/abs-2006-01410,DBLP:conf/accv/FajtlSAMR18,DBLP:journals/corr/abs-1708-09545,DBLP:conf/mm/FengLKZ18}. 

In this paper, we address this research gap and investigate how different feature types, i.e., static and motion features, can be integrated in a model architecture for video summarization.
How a fusion of different types of features affects the overall performance by incorporating them with an attention mechanism similar as used by previous work~\cite{DBLP:journals/corr/abs-2006-01410, DBLP:conf/accv/FajtlSAMR18, DBLP:journals/corr/abs-1708-09545,DBLP:conf/wacv/FuTC19}. 
For this task, we propose a novel deep learning model for supervised video summarization called Multi-Source Visual Attention (MSVA). The model fuses image and motion features based on a self-attention mechanism~\cite{DBLP:conf/accv/FajtlSAMR18} in a parallel fashion. Our comprehensive analysis on two benchmark datasets shows that our model outperforms other systems on the \emph{SumMe} dataset~\cite{DBLP:conf/eccv/ZhangCSG16}, while obtaining performance similar to the state of the art on the \emph{TVSum} dataset~\cite{DBLP:conf/cvpr/SongVSJ15}. 
In addition, we uncover issues in the experimental evaluation of previous methods: in some cases, videos were either excluded from the evaluation or reused in multiple splits, which makes it difficult to compare the systems in a fair manner and hinders the reproducibility of results.
The crucial aspect of cross-fold validation on both benchmark datasets, where previous methods either excluded some data points from evaluation or some data points were repeated in multiple splits, which makes it difficult to compare the published systems fairly. 
Therefore, we present a revised version of both benchmark datasets by providing five new non-overlapping splits and evaluating previous approaches on them and share the source code of our methodology and the evaluation\footnote{https://github.com/TIBHannover/MSVA}. 
We share the source code for the proposed model and the new non-overlapping splits for the \emph{TVSum} and \emph{SumMe} datasets with the research community. 
Our main contributions can be summarized as follows:

1.)  We introduce a novel architecture based on multi-source visual features with an attention mechanism.
Track changes is off

2.) We identify issues in previous experimental setups and reproduce the experimental results for some approaches on valid cross-validation folds for two benchmark datasets.

3.) State-of-the-art results are improved for the \emph{SumMe} dataset, while achieving similar results in comparison with other models on the \emph{TVSum} dataset.

The rest of the paper is structured as follows. In Section~\ref{sec:related_work}, we review previous work on supervised video summarization. Section~\ref{sec:methodology} describes the different feature sets, the proposed model architecture, and the attention mechanism. Experimental results and the comparison with other state-of-the-art methods are reported in Section~\ref{sec:experiments}. We conclude the paper with a summary in Section~\ref{sec:conclusion}.

\begin{figure*}[ht]
\begin{center}
\includegraphics[width=1.0\textwidth,height=7cm]{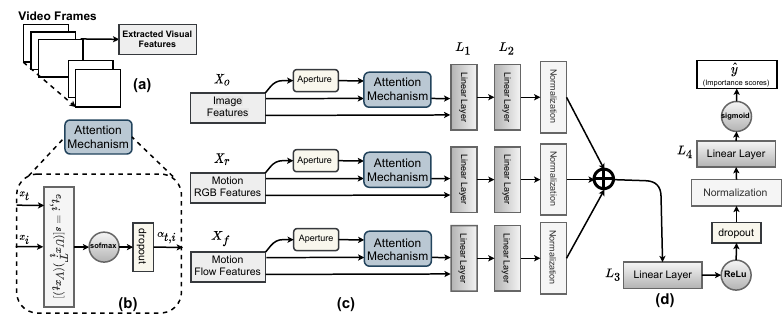}
\caption{The neural network architecture for the Multi-Source Visual Attention (MSVA) model with parallel self-attention mechanism based on multiple feature sets}
\label{fig:overll_model_architecture}
\end{center}
\end{figure*}
																																																																																																																																																																																																																			   
\section{Related Work}\label{sec:related_work}

To solve the task of video summarization, both supervised and unsupervised machine learning approaches have been suggested in the literature. 
Supervised methods train classifiers to learn the importance of a frame or segment for a summary. The process starts with the segmentation of videos,  either uniformly into equally sized chunks, as done by Gygli et al.~\cite{DBLP:conf/eccv/GygliGRG14}, or using algorithms like kernel temporal segmentation (KTS) by Potapov et al.~\cite{DBLP:conf/eccv/PotapovDHS14}. Gygli et al.~\cite{DBLP:conf/eccv/GygliGRG14} computed an \textit{interestingness} score for each segment using a weighted sum of features by combining low-level spatio-temporal salience or high-level motion information, while Song et al.~\cite{DBLP:conf/cvpr/SongVSJ15} measure frame-level importance using learned factorization. Another approach is suggested by Potapov et al.~\cite{DBLP:conf/eccv/PotapovDHS14} to train SVMs to classify frames in segments obtained through KTS.

Recurrent Neural Networks (RNN) or specifically long short-term memory (LSTM) and bidirectional LSTM (BiLSTM) have been also proposed for video summarization, where a BiLSTM model is stacked with Determinantal Point Process (DPP)~\cite{DBLP:conf/eccv/ZhangCSG16}, weighted memory layers with LSTM~\cite{DBLP:conf/eccv/ZhangGS18}.
In these approaches, either model helps to avoid similar frames in the final selection of a summary or solves this problem by encoding long video sequences to short sequences. As mentioned before, attention mechanisms are widely used in video summarization and combined with different neural architectures~\cite{DBLP:journals/corr/abs-2006-01410, DBLP:conf/accv/FajtlSAMR18,DBLP:journals/corr/abs-1708-09545,DBLP:conf/wacv/FuTC19} where promising or even best results have been achieved recently.

Our model \emph{MSVA} differs from approaches like MAVS~\cite{DBLP:conf/mm/FengLKZ18}, M-AVS ~\cite{DBLP:journals/corr/abs-1708-09545} and MC-VSA ~\cite{DBLP:journals/corr/abs-2006-01410} as follows. The proposed \emph{MSVA} model has multiple sources of visual features where attention is applied to each source in a parallel fashion. The MAVS system~\cite{DBLP:conf/mm/FengLKZ18} is a memory augmented video summarizer with global attention, M-AVS~\cite{DBLP:journals/corr/abs-1708-09545} considers multiplicative attention for video summarization with encoder-decoder, and MC-VSA~\cite{DBLP:journals/corr/abs-2006-01410} a multi-concept video self-attention where attention is applied to multiple layers of encoder and decoder. 
The majority of previous work~\cite{DBLP:conf/eccv/ZhangCSG16, DBLP:conf/eccv/ZhangGS18, DBLP:journals/corr/abs-1708-09545} use pre-trained image features from GoogleNet~\cite{DBLP:conf/cvpr/SzegedyLJSRAEVR15} to encode video frames.


\section{Supervised Video Summarization with Multi-source Features}\label{sec:methodology}

In this section, we describe the overall architecture of the Multi-Source Visual Attention (MSVA) model and details about the different building blocks. 
First, we define some variables of the target problem. A video can be represented as a sequence $V=\left(v_1,v_2,\ldots,v_t,\ldots,v_T\right)$, where $t \in {1,2,\dots,T}$ and $v_t$ is the visual frame at time $t$. 
The sequence $V$ of frames can be represented by different visual features as $X=\left(\mathbf{x_1},\mathbf{x_2},\ldots,\mathbf{x_t},\ldots,\mathbf{x_T}\right)$, where $\mathbf{x_t}  \in \mathbb{R}^{d}$ is a vector that represents the extracted features from the \emph{t-th} frame based on a specific feature encoder with a dimension $d$. The task of the model is to produce $Y=\{\mathbf{y_1},\mathbf{y_2},\ldots,\mathbf{y_t},\ldots,\mathbf{y_T}\}$ as an output that represents the importance score of frames.

The feature encoders can be based on pre-trained models like GoogleNet~\cite{DBLP:conf/cvpr/SzegedyLJSRAEVR15}, or content-based image features as mentioned in related work, in order to extract features to represent frames in videos. In contrast to prior work, we exploit additional models to enhance the representation of visual information in frames. For instance, different actions such as \textit{bungee jumping} or \textit{hiking} need to have different representations and rely on motion and temporal aspects, and not just on the static image content and object categories present in the ImageNet dataset. Therefore, we propose the use of content-based image features in combination with motion-related features to capture a richer representation within our model.

Once the features are extracted, we employ an attention mechanism followed by a couple of linear layers with different features and fuse them to obtain a common embedding space to represent frames. After fusion, we apply linear layers, normalization, activation functions, and predict the importance score of given input frames. The overall architecture of the \textit{MSVA} model is shown in Figure~\ref{fig:overll_model_architecture}. Next, we describe visual feature extraction, the attention mechanism, and fusion techniques to incorporate visual features from multiple sources for video summarization.

\subsection{Content-based Image and Motion Features}

We propose to combine three different feature types to provide a richer representation of frames in videos. After applying uniform sub-sampling to all frames of a video (two frames per second), the selected frames are passed as input to the pre-trained models to extract the following features.

\noindent
\textbf{Image content}: A deep neural network is trained on the ImageNet dataset for an object classification task ~\cite{DBLP:conf/cvpr/SzegedyLJSRAEVR15}. The most common pre-trained model to extract visual features for the video summarization domain is GoogleNet~\cite{DBLP:conf/cvpr/SzegedyLJSRAEVR15}. We use the same model to extract content-based frame features from the \textit{pool5} layer (\num{1024} dimensions), and represent them as $X_o$.

\noindent
\textbf{Motion}: To leverage motion-related features, we use the pre-trained I3D (Inflated 3D ConvNet)~\cite{DBLP:conf/cvpr/CarreiraZ17} model on Kinetics dataset~\cite{kinetics_dataset}, which is composed of human actions such as \textit{drawing}, \textit{drinking}, \textit{laughing}, \textit{hugging}, \textit{opening present}. From this model, we can extract two types of features: RGB (red, green, blue) and optical flow. RGB features, denoted as $X_r$, capture the channel-wise color information with regard to scene changes, while the optical flow, denoted as $X_f$, represents the motion in consecutive frames. The features are extracted from the second last layer of the pre-trained I3D model (\num{1024} dimensions).

\subsection{Attention Module}
The attention module used in our architecture is based on Fajtl et al.'s approach~\cite{DBLP:conf/accv/FajtlSAMR18}. It is shown in Figure~\ref{fig:overll_model_architecture}~(b). Each type of feature is fed into a separate attention mechanism followed by two linear and single normalization layers. An important aspect of the attention mechanism is the aperture $p$ that defines the aperture window with a range [$t$\text{-}$p$, $t$\text{+}$p$]; the normalized attention vectors for a window are $(\mathbf{\alpha_{t-p}},\dots,\mathbf{\alpha_{t-1}},\mathbf{\alpha_{t}},\mathbf{\alpha_{t+1}},\dots,\mathbf{\alpha_{t+p}})$. The attention weights at time index $t$ with a subset from the feature set $X$ are calculated as follows.

\begin{gather}
e_{t,i} = s[(U\mathbf{x_{i}}^T) (V\mathbf{x_{t}})], 
  \text{    }
t \in [0,T\text{-}1],
\text{    }
i \in [0, T\text{-}1]
\end{gather}

Here, $T$ is the number of video frames, $U$ and $V$ are learnable weight matrices, $s\in[0,1]$ is a scale parameter that reduces the value of the dot product, $\mathbf{x_i}$ is the i-th feature vector from the entire input sequence.

\begin{gather}
\alpha_{t,i} = \frac{\text{exp}(e_{t,i})}{\sum_{k=0}^{T\text{-}1} \text{exp} (e_{t,k})}
\end{gather}

$\alpha_{t,i}$ is a pairwise attention weight of the input vector $x_t$ with respect to a vector from the entire input sequence $x_i$. The vector $\mathbf{\alpha_{t}}$ contains attention weights for the target frame at time $t$ based on the other vectors from the input sequence.
\begin{gather}
\alpha_{t}= (\mathbf{\alpha_{t-p,i}},\dots,\mathbf{\alpha_{t-1,i}},\mathbf{\alpha_{t,i}},\mathbf{\alpha_{t+1,i}},\dots,\mathbf{\alpha_{t+p,i}})
\end{gather}

The feature vector of a visual descriptor is multiplied with attention weights and fed into two linear layers ($L_1$, $L_2$) and then into a normalization layer to obtain a latent representation $Z$ of each type of features (as shown in Figure~\ref{fig:overll_model_architecture}~(c)): 

\begin{equation}
Z = \text{Norm}(L_2(L_1(\alpha_tX)))\\
\end{equation}



\subsection{Multi-source Fusion}

As mentioned above, our proposed model incorporates different types of features. In this stage, all latent representations $Z$ after attention and the following linear transformation layers are fused to obtain a single vector representation of input frames. We use an addition function to fuse three vectors $Z_o$, $Z_f$, $Z_r$ corresponding to the latent representations of input features $X_o$, $X_f$, $X_r$ (image, optical flow, RGB). The result is passed to a linear layer ($L_3$), followed by \emph{ReLU} activation, dropout, normalization and another linear layer ($L_4$). Finally, the output vector is fed into a sigmoid function that outputs importance scores $\hat{y}$ for the input frames:

\begin{gather}
\label{eq:fusion}
 h = L_4(\text{Norm}(\text{Drop}(\text{ReLU}(L_3(Z_o + Z_f + Z_r)))))\\
 \hat{y} = sigmoid(h)
\end{gather}

The formula given in Equation~\ref{eq:fusion} and the model architecture depicted in Figure~\ref{fig:overll_model_architecture} can be considered as an \textit{intermediate} fusion, since the combination of latent representations of different feature types is performed in the intermediate layers of the neural network. Additionally, we experiment with other techniques such as \textit{early} and \textit{late} fusion. \textit{Early fusion} is realized by combining the input features $X$ after the encoding of input frames, then it is followed by a single attention mechanism, linear layers, normalization, dropouts, activation functions, and the classification layer. For the \textit{late fusion}, we combine the latent representations in the last linear layer ($L_4$), which indicates that latent representations of different types of features are processed by different layers in parallel until $L_4$.

\section{Experimental Setup and Results}\label{sec:experiments}

In this section, we present details about benchmark datasets, evaluation protocols, ablation studies, comparison with state-of-the-art baselines, and video-wise qualitative analysis.

\subsection{Datasets and Evaluation Metrics}
We use the following benchmark datasets to evaluate our approach and compare it with state-of-the-art approaches:

1.) TVSum~\cite{DBLP:conf/cvpr/SongVSJ15}: 50 videos with the length of 2-10 minutes, annotated by 20 users.

2.) SumMe~\cite{DBLP:conf/eccv/ZhangCSG16}: \num{25} videos with the length of 1-6 minutes, annotated by 18 users.

The evaluation of previous approaches on these datasets is based on 5-fold cross-validation and the reported results are $F_1$ scores averaged across the five splits of the corresponding datasets. The splits for TVSum and SumMe datasets are provided by Zhang et al.~\cite{DBLP:conf/eccv/ZhangCSG16}. When reproducing the results of previous work, we have observed methodological issues in the evaluation for both benchmark datasets. Some videos are dropped from certain splits, and are never part of the validation set. For instance, "video\_5" and "video\_8" in SumMe as well as "video\_21" and "video\_28" in TVSum are not part of any validation split. In total, eight videos from SumMe and \num{19} videos from TVSum are dropped from certain splits and were not evaluated. Another issue is that some videos are repeated in multiple splits. To fix the mentioned problems and to provide a fair comparison for future research, we release new versions of the two benchmark datasets by providing five \textit{non-overlapping} splits where videos are equally divided across splits without any repetition or exclusion.

Regarding the evaluation, Otani et al.~\cite{DBLP:conf/cvpr/OtaniNRH19} argue that the $F_{1}$ score for the task of video summarization has certain limitations and proposed to measure the correlation between predicted and human-annotations.
In particular, they suggested Spearman’s $\rho$ and Kendall’s $\tau$ as correlation coefficients to evaluate models on how close the summaries predicted by models are to human annotations. Following these arguments, we evaluate our model architecture on the original splits and on the new \textit{non-overlapping} splits for both datasets in terms of $F_{1}$ scores and correlation coefficients. 
The corresponding non-overlapping splits and the source code for evaluation metrics will be available to enable fair comparisons and reproducibility of future research\footnote{https://github.com/TIBHannover/MSVA}.  
We have reproduced work from Fajtl et al.~\cite{DBLP:conf/accv/FajtlSAMR18} with correlation coefficients according to~\cite{DBLP:journals/corr/abs-2006-01410, DBLP:conf/cvpr/OtaniNRH19} for the respective experiments by evaluating on both $F_{1}$ measure and correlation coefficients to compare against previous methods.

\subsection{Results}

\begin{table}[!ht]
\caption{Ablation study with different feature types, fusion techniques with best aperture size (250) for the \textit{MSVA} model. $F_{1}$ is the average score calculated on the newly provided five non-overlapping splits.}
\centering
\begin{tabular}{|*{4}{l|}}
\hline 
\textbf{Dataset}  & Fusion & Features& $F_{1}$ \\ \hline
  & - & $X_o$ & 50.5 \\  \cline{2-4} 
  & \multirow{3}{5em}{early} & $X_{o}$+$X_{r}$+$X_{f}$ & 46.7 \\ \cline{3-4} 
  &  & $X_{o}$+$X_{r}$ &  44.5 \\ \cline{3-4} 
\multirow{4}{4em}{SumMe}   &  & $X_{o}$+$X_{f}$ &  44.8 \\ \cline{2-4} 
   
 &  \multirow{3}{5em}{intermediate} & $X_o$+$X_r$+$X_f$ & \textbf{53.4} \\ \cline{3-4}
 &  & $X_{o}$+$X_{r}$ & 50.9 \\ \cline{3-4} 
&  & $X_{o}$+$X_{f}$ &  51.5 \\ \cline{2-4} 
   
 & \multirow{3}{5em}{late}  & $X_o$+$X_r$+$X_f$ & 51.0 \\  \cline{3-4}
 &  & $X_{o}$+$X_{r}$  & 50.1 \\ \cline{3-4} 
&  & $X_{o}$+$X_{f}$  & 50.8 \\ \cline{2-4}

\hline 
\hline 
 
 & - & $X_o$  & 60.1 \\ \cline{2-4}
 & \multirow{3}{5em}{early}  & $X_o$+$X_r$+$X_f$  & 57.3 \\  \cline{3-4}
&  & $X_{o}$+$X_{r}$ & 56.7 \\ \cline{3-4} 
\multirow{4}{4em}{TVSum} &  & $X_{o}$+$X_{f}$ &  56.3 \\ \cline{2-4} 
   
 & \multirow{3}{5em}{intermediate} & $X_o$+$X_r$+$X_f$ &  \textbf{61.5} \\  \cline{3-4}
 &  & $X_{o}$+$X_{r}$ & 61.1 \\ \cline{3-4} 
&  & $X_{o}$+$X_{f}$ & 61.2 \\ \cline{2-4}

 & \multirow{3}{5em}{late} & $X_o$+$X_r$+$X_f$ & 60.1 \\  \cline{3-4}
 &  & $X_{o}$+$X_{r}$ &  58.9 \\ \cline{3-4} 
&  & $X_{o}$+$X_{f}$ &  59.7 \\ \cline{2-4} 
\hline 
 
\end{tabular}
\label{tab:ablation}
\end{table}


\textbf{Ablation study}: We evaluated different building blocks of our proposed \textit{MSVA} model. We performed a grid search on model hyper-parameters such as the aperture size of an attention mechanism (\num{50}-\num{1000}), fusion techniques (early, intermediate, late), and combination of feature types (object, RGB, optical flow) using random \num{20}\% of data. 
The linear layers are of size \num{1024}, \emph{Adam} (Adaptive Moment Estimation) is used as the optimizer, and each variation is trained for \num{300} epochs with a stopping criterion of 50 epochs when the loss is static.

The results are given in Table~\ref{tab:ablation}. 
The reported score is the average $F_1$ of 5-fold cross-validation on the \textit{non-overlapping} splits for both benchmark datasets that we provide. We include only the best performing combinations of hyper-parameters (including aperture window size 250). It can be concluded that the model with an intermediate fusion of three features gives the best performance for both datasets with an aperture size of \num{250}.

\textbf{Overall comparison}: We compared our proposed \textit{MSVA} model with other state-of-the-art models that were evaluated on both benchmark datasets. 
																				
The results are given in  Table~\ref{tab:overallcompare} for both benchmark datasets. We report $F_1$ score, Kendall’s $\tau$ and Spearman’s $\rho$ correlation coefficients on the newly provided \textit{non-overlapping} splits (denoted as $F_1$) along with $F_1$ on the original splits (denoted as $F_{1}^*$). The results of previous work that did not share source code are reported only for the original splits. We evaluated VASNet~\cite{DBLP:conf/accv/FajtlSAMR18} on all evaluation metrics as the source code of the model is available\footnote{https://github.com/ok1zjf/VASNet}. Based on the given results, we can see that our model improves the state-of-the-art results for \emph{SumMe} dataset and achieves comparable results for the \emph{TVSum} dataset, while outperforming multiple systems including VASNet~\cite{DBLP:conf/accv/FajtlSAMR18}. A similar pattern is seen for both correlation coefficients, where our model obtains the best results for both datasets. 
Another observation is that the performance of VASNet is reduced by 1-2 points in terms of $F_1$ score when evaluated on \textit{non-overlapping} splits in comparison to the original splits of both datasets. It can be explained by the fact that some videos were not part of any splits for 5-fold cross-validation and some videos were repeated across multiple splits. The MAVS approach~\cite{DBLP:conf/mm/FengLKZ18} is the best performing model on the \emph{TVSum} from the compared models, while it has poor performance on \emph{SumMe}. On the contrary, our proposed model outperforms all baselines on \emph{SumMe}, while still achieving comparable results for \emph{TVSum} as well, and best results in terms of correlations.

\begin{table}[ht]
\caption{Comparison of different methods on benchmark datasets. $F_{1}$is an average score calculated on the newly provided five non-overlapping splits, $F_{1}^*$ is the reported score by previously systems on original five splits. $\rho$ is Spearman’s and $\tau$ is Kendall’s correlation coefficients.}
\centering
\begin{tabular}{|*{6}{l|}}
\hline

\textbf{Dataset} & Method & $F_{1}^*$ & $F_{1}$ &  $\tau$ &  $\rho$ \\ \hline

\multirow{6}{2.2em}{SumMe} & MAVS \cite{DBLP:conf/mm/FengLKZ18} & 43.1 & - & - & - \\ \cline{2-6}
 & M-AVS \cite{DBLP:journals/corr/abs-1708-09545} & 44.4 & - & - & - \\ \cline{2-6} 
&  re-SEQ2SEQ \cite{DBLP:conf/eccv/ZhangGS18} & 44.9 & - & - & -\\ \cline{2-6} 
&  VASNet \cite{DBLP:conf/accv/FajtlSAMR18} & 49.7 & 48.0 & 0.16 & 0.17
\\
\cline{2-6} &  MC-VSA \cite{DBLP:journals/corr/abs-2006-01410} & 51.6 & - & - & - \\
\cline{2-6} & MSVA \textbf{(ours)} & \textbf{54.5}  & \textbf{53.4} & \textbf{0.20} & \textbf{0.23} \\

 \hline \hline 

 \multirow{6}{2.2em}{TVSum} & M-AVS \cite{DBLP:journals/corr/abs-1708-09545} &61.0 & - & - & - \\  \cline{2-6}
 &  VASNet \cite{DBLP:conf/accv/FajtlSAMR18} &61.4 & 59.8  & 0.16 & 0.17 \\  \cline{2-6}
 &  MC-VSA \cite{DBLP:journals/corr/abs-2006-01410} & 63.7 & - & 0.116 & 0.142 \\  \cline{2-6} 
 &  re-SEQ2SEQ \cite{DBLP:conf/eccv/ZhangGS18} &63.9 & - & - & - \\ \cline{2-6}
 & MAVS \cite{DBLP:conf/mm/FengLKZ18} & \textbf{67.5} & - & - & - \\  \cline{2-6}
 & MSVA \textbf{(ours)} &62.8 & \textbf{61.5} & \textbf{0.19} & \textbf{0.21} \\
\hline 
 
\end{tabular}%

\label{tab:overallcompare}
\end{table}

\begin{figure*}[!ht]
\begin{center}
\includegraphics[width=0.9\textwidth,height=5cm]{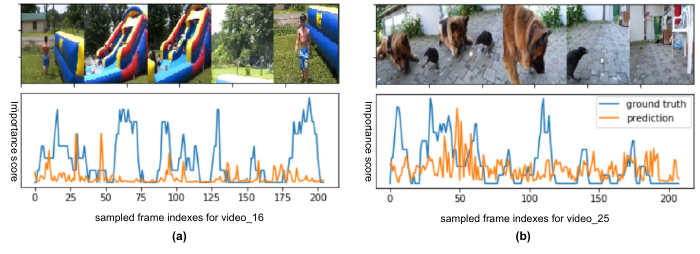}
\caption{Comparison of predictions and ground truth labels on videos with low (a) ``Playing on water slide'' and high (b) ``Playing ball'' score from the SumMe dataset}
\label{fig:example_good_bad}
\end{center}
\end{figure*}

\subsection{Qualitative analysis}

In Figure~\ref{fig:example_good_bad}, we plot the importance score predictions of \textit{MSVA} model compared to the average ground truth labels assigned by the annotators. Our model achieved lower performance for the video on the left, while it achieved a higher performance for the video on the right. The video on the left is called ``Playing on water slide'' with main content being recorded next to a water slide and a number of kids are playing around it. The difficulty lies in selecting different importance scores for frames that look similar, i.e., the background is still a water slide and children. Such confusion can be observed for ground truth labels assigned in the middle of the video where different scores are assigned to visually similar frames. 
The video on the right is called ``Playing ball'' with the main content being a dog and a bird playing with a white ball. One analysis in this video is that there are few objects appearing at a particular time, i.e., three objects, where one is prominent and the others are small like a bird and a ball.

\begin{figure}[ht]
\begin{center}
\includegraphics[width=0.5\textwidth,height=6cm]{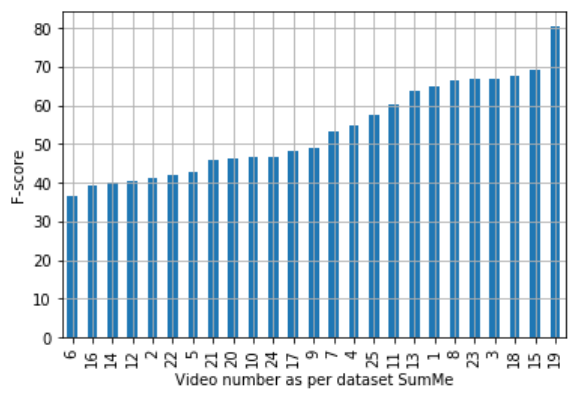}
\caption{$F_{1}$ score analysis for all videos in SumMe. The scores are taken when videos were used for testing across 5-folds.}
\label{fig:video_wise-analysis}
\end{center}
\end{figure}

To understand the impact of splitting the dataset, we plot $F_{1}$ scores for all videos in SumMe in Figure~\ref{fig:video_wise-analysis}. The scores are computed when an individual video was part of the test set across five folds. There is a big difference in the F1 score across videos. Thus, the exclusion of certain videos affects the overall comparison across different models as well as the repetition of the same videos in multiple splits has an impact.

\subsection{Discussion}
The experimental results for both datasets demonstrate the importance of exploiting multiple sources of feature types, particularly with regard to motion. Similarly, the attention mechanism on each source in a parallel fashion plays a vital role to provide decisive power to the model. Although the supervised method has the advantage to learn from annotated labels, the model also incorporates bias from datasets based on (partially disagreeing) annotations from multiple users, as shown by the qualitative analysis mentioned above. Lastly, it is shown that the fusion strategy has a noticeable impact on the performance when utilizing multiple feature sets. 
																																																																																																																																																																					
 Model performance varies a lot on different videos for which the reasons ca reasons can be given by observing the visual content of the video. This also demands to get features from the visual contextual domain to fill the gap on these difficult kinds of videos.

\section{Conclusion}\label{sec:conclusion}

In this paper, we have proposed a model architecture that utilizes visual features from multiple sources, i.e., static object and motion features, with an attention mechanism and different fusion techniques. The intermediate fusion of object and motion features appeared to be the best configuration as shown by experiment results. State-of-the-art results were improved on the benchmark dataset SumMe and comparable results were obtained on the other benchmark dataset TVSum. Furthermore, methodological issues have been identified in the evaluation setup of previous work, and we have provided non-overlapping splits for cross-validation for a fair comparison. In the future, we will focus on semantic aspects to enhance the model with additional decisive capabilities.

\bibliographystyle{IEEEbib}
\bibliography{references}

\begin{thebibliography}{10}

\bibitem{DBLP:conf/cvpr/SongVSJ15}
Yale Song, Jordi Vallmitjana, Amanda Stent, and Alejandro Jaimes,
\newblock ``{TVSum}: Summarizing web videos using titles,''
\newblock in {\em {IEEE} Conference on Computer Vision and Pattern Recognition,
  {CVPR}}. 2015, pp. 5179--5187, {IEEE} Computer Society.

\bibitem{DBLP:conf/eccv/ZhangCSG16}
Ke~Zhang, Wei{-}Lun Chao, Fei Sha, and Kristen Grauman,
\newblock ``Video summarization with long short-term memory,''
\newblock in {\em {ECCV} - 14th European Conference on Computer Vision}. 2016,
  vol. 9911, pp. 766--782, Springer.

\bibitem{DBLP:conf/eccv/ZhangGS18}
Ke~Zhang, Kristen Grauman, and Fei Sha,
\newblock ``Retrospective encoders for video summarization,''
\newblock in {\em {ECCV} - 15th European Conference on Computer Vision}. 2018,
  vol. 11212, pp. 391--408, Springer.

\bibitem{DBLP:conf/eccv/PotapovDHS14}
Danila Potapov, Matthijs Douze, Za{\"{\i}}d Harchaoui, and Cordelia Schmid,
\newblock ``Category-specific video summarization,''
\newblock in {\em {ECCV} - 13th European Conference on Computer Vision}. 2014,
  vol. 8694, pp. 540--555, Springer.

\bibitem{DBLP:conf/eccv/GygliGRG14}
Michael Gygli, Helmut Grabner, Hayko Riemenschneider, and Luc~Van Gool,
\newblock ``Creating summaries from user videos,''
\newblock in {\em {ECCV} - 13th European Conference on Computer Vision}. 2014,
  vol. 8695, pp. 505--520, Springer.

\bibitem{DBLP:conf/mm/Wang000FT19}
Junbo Wang, Wei Wang, Zhiyong Wang, Liang Wang, Dagan Feng, and Tieniu Tan,
\newblock ``Stacked memory network for video summarization,''
\newblock in {\em Proceedings of the 27th {ACM} International Conference on
  Multimedia, {MM}}. 2019, pp. 836--844, {ACM}.

\bibitem{DBLP:conf/nips/ShiCWYWW15}
Xingjian Shi, Zhourong Chen, Hao Wang, Dit{-}Yan Yeung, Wai{-}Kin Wong, and
  Wang{-}chun Woo,
\newblock ``Convolutional {LSTM} network: {A} machine learning approach for
  precipitation nowcasting,''
\newblock in {\em Advances in Neural Information Processing Systems 28: Annual
  Conference on Neural Information Processing Systems 2015, December 7-12,
  2015, Montreal, Quebec, Canada}, 2015, pp. 802--810.

\bibitem{DBLP:conf/nips/GongCGS14}
Boqing Gong, Wei{-}Lun Chao, Kristen Grauman, and Fei Sha,
\newblock ``Diverse sequential subset selection for supervised video
  summarization,''
\newblock in {\em Annual Conference on Neural Information Processing Systems},
  2014, pp. 2069--2077.

\bibitem{DBLP:journals/corr/abs-2006-01410}
Yen{-}Ting Liu, Yu{-}Jhe Li, and Yu{-}Chiang~Frank Wang,
\newblock ``Transforming multi-concept attention into video summarization,''
\newblock in {\em {ACCV} - 15th Asian Conference on Computer Vision}. 2020,
  vol. 12626, pp. 498--513, Springer.

\bibitem{DBLP:conf/accv/FajtlSAMR18}
Jiri Fajtl, Hajar~Sadeghi Sokeh, and Vasileios~Argyriou et~al.,
\newblock ``Summarizing videos with attention,''
\newblock in {\em {ACCV} Workshops - 14th Asian Conference on Computer Vision}.
  2018, vol. 11367, pp. 39--54, Springer.

\bibitem{DBLP:journals/corr/abs-1708-09545}
Zhong Ji, Kailin Xiong, Yanwei Pang, and Xuelong Li,
\newblock ``Video summarization with attention-based encoder-decoder
  networks,''
\newblock {\em {IEEE} Trans. Circuits Syst. Video Technol.}, vol. 30, no. 6,
  pp. 1709--1717, 2020.

\bibitem{DBLP:conf/mm/FengLKZ18}
Litong Feng, Ziyin Li, and Zhanghui~Kuang et~al.,
\newblock ``Extractive video summarizer with memory augmented neural
  networks,''
\newblock in {\em {ACM} Multimedia Conference on Multimedia Conference, {MM}}.
  2018, pp. 976--983, {ACM}.

\bibitem{DBLP:conf/wacv/FuTC19}
Tsu-Jui Fu, Shao-Heng Tai, and Hwann-Tzong Chen,
\newblock ``Attentive and adversarial learning for video summarization,''
\newblock {\em IEEE Winter Conference on Applications of Computer Vision
  (WACV)}, pp. 1579--1587, 2019.

\bibitem{DBLP:conf/cvpr/SzegedyLJSRAEVR15}
Christian Szegedy, Wei Liu, and Yangqing~Jia et~al.,
\newblock ``Going deeper with convolutions,''
\newblock in {\em {IEEE} Conference on Computer Vision and Pattern Recognition,
  {CVPR}}. 2015, pp. 1--9, {IEEE} Computer Society.

\bibitem{DBLP:conf/cvpr/CarreiraZ17}
Jo{\~{a}}o Carreira and Andrew Zisserman,
\newblock ``Quo vadis, action recognition? {A} new model and the kinetics
  dataset,''
\newblock in {\em 2017 {IEEE} Conference on Computer Vision and Pattern
  Recognition, {CVPR}}. 2017, pp. 4724--4733, {IEEE} Computer Society.

\bibitem{kinetics_dataset}
Will Kay, Brian~Zhang Jo{\~{a}}o~Carreira, Karen~Simonyan, and Chloe Hillier,
\newblock ``The kinetics human action video dataset,''
\newblock {\em CoRR}, vol. abs/1705.06950, 2017.

\bibitem{DBLP:conf/cvpr/OtaniNRH19}
Mayu Otani, Yuta Nakashima, Esa Rahtu, and Janne Heikkil{\"{a}},
\newblock ``Rethinking the evaluation of video summaries,''
\newblock in {\em {IEEE} Conference on Computer Vision and Pattern Recognition,
  {CVPR}}. 2019, pp. 7596--7604, Computer Vision Foundation / {IEEE}.

\end{thebibliography}

\end{document}